\documentclass[10pt,letterpaper]{article}
\usepackage[T1]{fontenc}
\usepackage[nofontspec]{newtxtext}
\usepackage{steward}
\usepackage{amsmath,amssymb,graphicx,booktabs,array}
\usepackage{microtype}
\usepackage{enumitem}
\usepackage{caption}
\usepackage{flafter}
\usepackage[pdfversion=1.7,hidelinks]{hyperref}
\hypersetup{
 pdftitle={Skill Adapter: Personalized Modification and Online Routing of General-Purpose Skills},
 pdfauthor={Qijia Zhuang, Yanyin Gou, Wenbin Ma, Yuxiang Ren},
 pdfsubject={English manuscript in the original Skill Steward layout}}
\setlist[itemize]{leftmargin=1em,itemsep=0pt,parsep=0pt,topsep=3pt}
\title{Skill Adapter: Personalized Modification\\and Online Routing of General-Purpose Skills}
\author{Qi{}jia Zhuang\textsuperscript{1,*},\quad
Yanyin Gou\textsuperscript{1,*},\quad
Wenbin Ma\textsuperscript{2,*},\quad
Yuxiang Ren\textsuperscript{1,\(\dagger\)}}
\affiliation{\textsuperscript{1}Nanjing University\qquad
\textsuperscript{2}Jiangnan University}
\authornote{\textsuperscript{*}These authors contributed equally and share first authorship.
\quad\textsuperscript{\(\dagger\)}Corresponding author.}
\date{}
\begin{document}
\maketitle
\begin{abstract}
Large language model (LLM) agents use external skills to reuse task procedures and experience with tools. General-purpose skills describe shared execution procedures, while specific tool choices, operating practices, and output requirements vary across users. In long-horizon tasks, subgoals, available inputs, and environmental states change as execution progresses. Agents must select currently applicable skills from an expanding skill pool and prepare candidates for subsequent steps.

We propose Skill Adapter, which enables personalized modification and online routing of general-purpose skills. The system distills local adjustments from user requirements and execution feedback and stores them as Adapters linked to general-purpose skills. The routing agent selects general-purpose skills and applicable Adapters based on the current subtask, execution progress, and environmental feedback, incrementally maintaining a task-level online skill library that the main LLM reads on demand. The system preloads candidates for subsequent nodes and rechecks their applicability once actual inputs are determined. Reinforcement learning optimizes the two types of decisions using modification gains and routing outcomes, respectively.

We evaluate Skill Adapter on ALFWorld, WebShop, and SkillsBench. Compared with SkillRL, it improves task success rates by an average of \textbf{11.43 percentage points} and reduces mean execution time on all three benchmarks. Skill Adapter retains its performance advantage when the skill pool grows to 1,000 skills.

\end{abstract}
\section{Introduction}
\label{sec:introduction}

Large language model (LLM) agents are increasingly capable of planning, invoking tools, and interacting continuously with their environments. Their applications have expanded from single-turn question answering to web interaction, embodied decision-making, and complex workflows (Yao et al.\ 2023; Schick et al.\ 2023; Liu et al.\ 2023; Wu et al.\ 2023). In these tasks, organizing past experience into reusable skills helps reduce repeated exploration and improve the stability of long-horizon execution (Wang et al.\ 2023; Zhao et al.\ 2023; Shinn et al.\ 2023). Skill pools thus provide an external capability layer that connects task experience to agent execution.

General-purpose skills describe shared procedures for a class of tasks, while specific execution procedures vary with user requirements. For example, when translating an English contract into a Chinese PDF, some users focus on the translated content, while others need to preserve the original pagination, layout, and annotations. Both sets of requirements involve document extraction, translation, and PDF generation, but differ in the fields to extract, tool parameters, and checks to perform.

Personalized modification must express user-specific operating requirements within shared procedures. Online routing must select the general-purpose skills and applicable Adapters for the current subtask according to the execution state. Semantic matching between tasks and skills provides initial candidates, while actual inputs, tool availability, and execution feedback determine whether each candidate can be used at the current step.

\begin{figure}[t]
  \centering
  \includegraphics[width=\columnwidth]{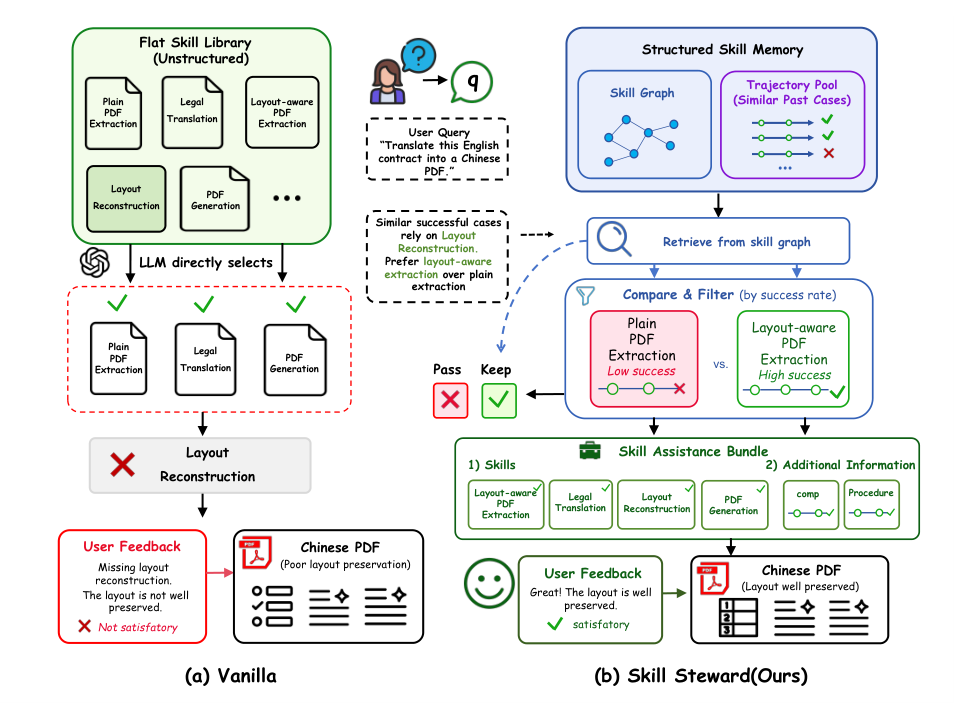}
  \caption{The influence of user requirements on skill selection and composition. In contract translation, layout-preservation requirements affect document extraction and make layout reconstruction part of the subsequent workflow. Task descriptions and user feedback jointly inform skill adjustments and the selection of execution procedures.}
  \label{fig:motivation}
\end{figure}

Subsequent steps in long-horizon tasks depend on the actual results of preceding operations. For example, contract layout reconstruction requires selecting a procedure based on the extracted layout information, and feedback from checking the generated output may introduce further processing requirements. Routing must therefore update skill provision as execution progresses, reusing valid entries and adding or replacing candidates. Growth of the global skill pool also increases the cost of retrieval and selection.

To address these challenges, we propose \textbf{Skill Adapter}. The framework uses general-purpose skills as a shared foundation and stores user-specific modifications as Adapters linked to their original skills. Each Adapter describes applicability conditions and local adjustments to tools, parameters, operating steps, and output checks. As the task progresses, online routing selects general-purpose skills and their applicable Adapters and organizes the results into a task-level online skill library. The main LLM reads the skills needed for the current node on demand, while the routing agent prepares candidates for subsequent nodes and incrementally updates library entries in response to new inputs and feedback.

We use group relative policy optimization (GRPO) to train the modification and routing policies. The modification policy receives rewards based on changes in task success rate, while the routing policy receives rewards based on task execution outcomes and the quality of skill selection. Our contributions are threefold:

\begin{itemize}
  \item We introduce a personalized modification mechanism for general-purpose skills that expresses user requirements through local Adapters and preserves their links to original skills and supporting feedback.
  \item We introduce an online routing mechanism for long-horizon tasks that selects general-purpose skills and their Adapters according to the execution state, supporting continued execution through incremental updates to the online skill library and preloading for subsequent nodes.
  \item On ALFWorld, WebShop, and SkillsBench, Skill Adapter achieves higher task success rates than all compared baselines. Compared with SkillRL, it improves success rates by an average of \textbf{11.43 percentage points} and reduces mean execution time on all three benchmarks. Skill Adapter retains a success-rate advantage in the scaling experiment with 1,000 skills.
\end{itemize}

\section{Related Work}
\label{sec:related-work}

\subsection{Skill Extraction and Reuse}

Voyager generates executable functions from successful trajectories and maintains an expanding skill pool (Wang et al.\ 2023). ExpeL derives experiential rules from successful and failed trajectories, while Reflexion uses verbal feedback to improve subsequent decisions (Zhao et al.\ 2023; Shinn et al.\ 2023). Skill Adapter distills historical execution experience into general-purpose skills and uses associated Adapters to express the local modifications required by different users.

\subsection{Skill Retrieval and Routing}

SkillsBench evaluates the effects of external skills on task execution, while Skills-in-the-Wild studies retrieval and use within large skill collections (Li et al.\ 2026b; Liu et al.\ 2026). Graph of Skills uses explicit skill relations to support structured retrieval, SkillDAG organizes dependencies, conflicts, and specialization information through typed relations, and SkillRouter selects skills through retrieval and reranking (Li et al.\ 2026a; Bai et al.\ 2026; Zheng et al.\ 2026).

Skill Adapter focuses on online routing during long-horizon tasks. The system retrieves general-purpose skills based on the current subtask and execution state, then selects applicable Adapters under each skill. Routing results enter a task-level online skill library and are updated incrementally as inputs, environmental feedback, and task progress change. Preloading for subsequent nodes and reusing valid entries jointly support the continuous provision of skills.

\subsection{Reinforcement Learning and Skill Optimization}

Policy optimization methods such as PPO and GRPO support training language models with environmental feedback; GRPO constructs its optimization signal from relative rewards within a group (Schulman et al.\ 2017; Shao et al.\ 2024). Among skill-augmented agents, SAGE incorporates skill accumulation into sequential rollouts, while SkillRL constructs a hierarchical skill library from task experience and optimizes the skill-use policy (Wang et al.\ 2025; Xia et al.\ 2026). We apply reinforcement learning to personalized modification and online routing. The modification policy receives feedback from task performance before and after skill adjustments, while the routing policy receives feedback from execution outcomes and the quality of skill selection.

\section{Method}
\label{sec:method}

\begin{figure*}[t]
\centering
\includegraphics[width=\textwidth]{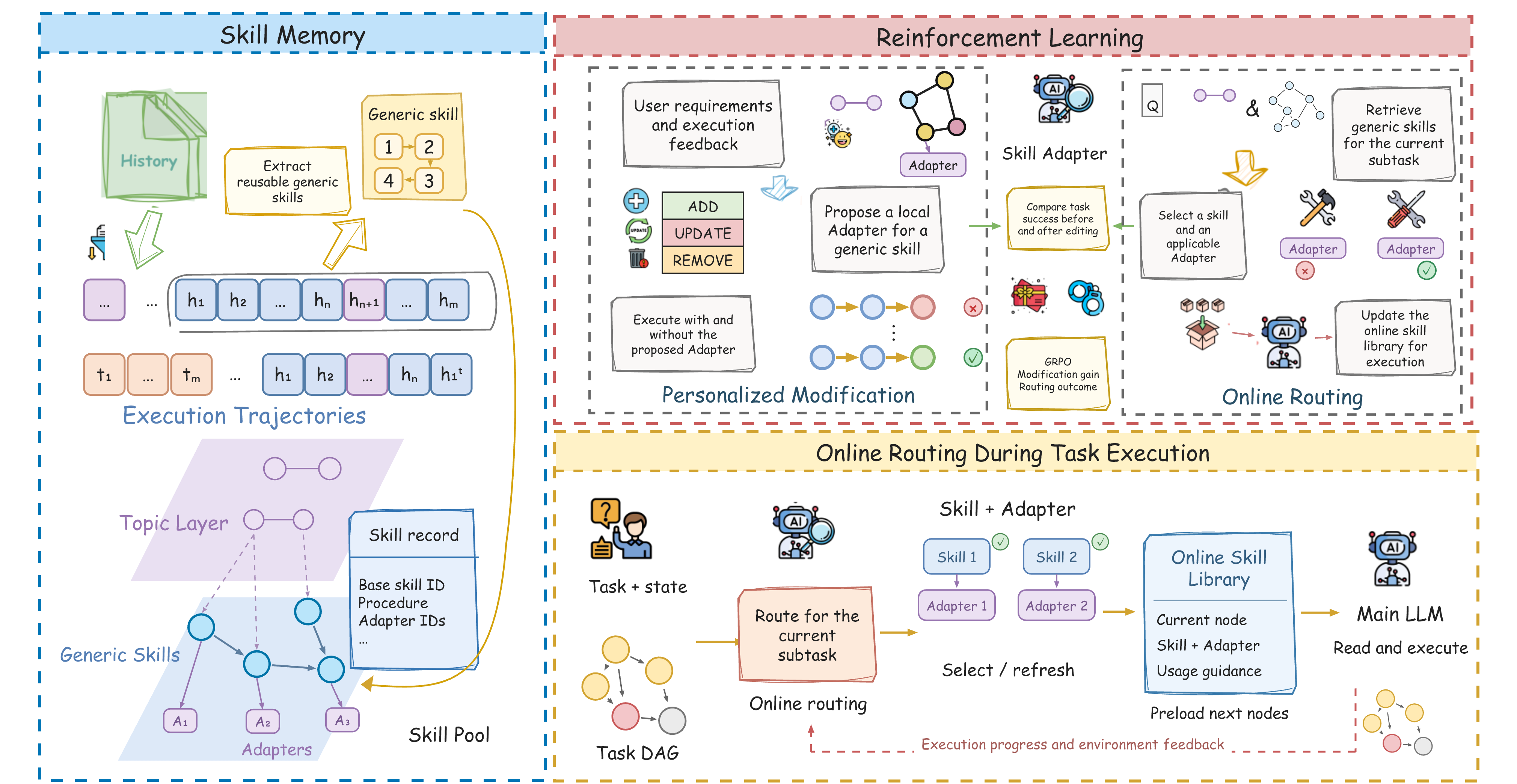}
\caption{The Skill Adapter framework. The system extracts general-purpose skills from historical trajectories and maintains the Adapters associated with each skill based on user requirements and execution feedback. During task execution, the routing agent selects general-purpose skills and applicable Adapters for the current DAG node, incrementally updates the online skill library, and preloads candidates for subsequent nodes. The main LLM reads skills and Adapters as needed, while execution progress and environment feedback inform the next routing round. GRPO optimizes the two policies using the gains from personalized modification and the execution outcomes of online routing, respectively.}
\label{fig:framework}
\end{figure*}

As shown in Figure~\ref{fig:framework}, Skill Adapter implements personalized modification and online routing over a shared pool of general-purpose skills, using reinforcement learning to optimize both types of decisions. Given a user $u$ and a task $q$, the system represents the user's explicitly stated preferences, operational requirements, and historical feedback through a personalization context $P_u$. At step $t$ of a long-horizon task, the available context is
\begin{equation}
x_{u,t}=\langle q,P_u,h_t\rangle,
\label{eq:context}
\end{equation}
where $h_t$ contains completed steps, current environment observations, and available intermediate results. The context $P_u$ combines the current request with previously obtained information and is updated as user feedback arrives. Task-specific requirements are used for the current task, while preferences that generalize across tasks are used for subsequent adaptation.

\subsection{Personalized Modification}

\subsubsection{General-Purpose Skills and Execution Evidence}

The system first constructs a general-purpose skill pool $\mathcal{G}$ by extracting reusable task objectives, operation sequences, tool invocation procedures, and input and output constraints from historical trajectories. Existing skills can also serve directly as the shared foundation. Each general-purpose skill is represented as
\begin{equation}
s_i=\langle n_i,d_i,c_i,I_i,O_i\rangle,
\label{eq:skill}
\end{equation}
where $n_i$ and $d_i$ denote its name and functional description, $c_i$ is its general execution procedure, and $I_i$ and $O_i$ specify input and output constraints. Skills are organized by topic and retain directed composition relations supported by successful trajectories, allowing candidate expansion during routing.

The trajectory pool records tasks, personalization contexts, the skills and versions actually invoked, environment observations, and final feedback. A trajectory for user $u$ is represented as
\begin{equation}
\tau_u=\langle q,P_u,\{e_t\}_{t=1}^{L_\tau},F,y\rangle,
\label{eq:trajectory}
\end{equation}
where $F$ denotes user feedback or environment diagnostics, and $y$ denotes the task outcome. Each execution step $e_t$ records the general-purpose skill ID, Adapter ID, environment action, and observation. The system distills general procedures that can be reused across tasks from trajectories and generates personalized modifications using execution evidence relevant to the current user's requirements.

\subsubsection{Personalized Adapters}

For a general-purpose skill $s_i$, an Adapter for user $u$ is defined as
\begin{equation}
\begin{aligned}
a_{u,i,k}&=\langle i,p_{u,i,k},\Delta_{u,i,k},\kappa_{u,i,k}\rangle,\\
\widetilde{s}_{u,i,k}&=\operatorname{Apply}(s_i,\Delta_{u,i,k}).
\end{aligned}
\label{eq:adapter}
\end{equation}
Here, $i$ identifies the associated general-purpose skill, and $k$ distinguishes the user's adapted versions for different conditions. The term $p_{u,i,k}$ describes the applicability conditions, $\Delta_{u,i,k}$ is a local modification to the skill's execution instructions, and $\kappa_{u,i,k}$ stores usage records and associated feedback evidence. At execution time, the local modification is applied to the general procedure to obtain the personalized skill $\widetilde{s}_{u,i,k}$.

Local modifications may concern tool selection, parameter settings, the addition or removal of steps, output formats, and verification requirements. For example, to meet a requirement to preserve the layout of a contract, an Adapter can add requirements to extract text positions, page numbers, and reading order to the general-purpose PDF extraction procedure.

Let $\mathcal{A}_u$ denote the set of Adapters for user $u$. Based on the general-purpose skill, personalization context, and relevant trajectories, the modification policy proposes an action to add, update, or delete an Adapter:
\begin{equation}
\begin{aligned}
m_t&\sim\pi_{\mathrm{mod}}\!\left(
\cdot\mid s_i,P_u,\mathcal{B}_{u,i},\mathcal{A}_u
\right),\\
\mathcal{A}_u'&=\operatorname{Update}(\mathcal{A}_u,m_t).
\end{aligned}
\label{eq:modification}
\end{equation}
Here, $\mathcal{B}_{u,i}$ contains historical records relevant to the user's requirements and skill $s_i$. The system distills reusable modifications from recurring deviations and applies them according to their applicability conditions. When the user's requirements change, the corresponding Adapter is updated based on feedback. Adapters and general-purpose skills are stored separately, and the same general-purpose skill can be associated with adapted versions for multiple users.

\subsection{Online Routing}

\subsubsection{Online Selection of Skills and Adapters}

Online routing continues throughout the execution of a long-horizon task. The main LLM decomposes the task into a directed acyclic graph (DAG) $\mathcal{D}$. The routing agent processes ready nodes according to the current execution progress and prepares candidate skills for subsequent nodes. For task node $q_j$ at step $t$, the available context is $x_{u,j,t}=\langle q_j,P_u,h_t\rangle$. Based on the node's objective and available intermediate results, the system retrieves a candidate set from the general-purpose skill pool:
\begin{equation}
\mathcal{R}_{j,t}=\operatorname{Retrieve}(q_j,h_t,\mathcal{G}).
\label{eq:retrieval}
\end{equation}

Retrieval uses topic-based search and semantic matching, followed by candidate expansion along skill composition relations. Each candidate skill is associated with its Adapters. The system filters these Adapters according to the current inputs, tool states, and applicability conditions to construct skill-version pairs:
\begin{equation}
\begin{aligned}
&\mathcal{C}_{u,j,t}
=\bigl\{(s_i,a):s_i\in\mathcal{R}_{j,t},\\
&\qquad a\in\mathcal{A}_{u,i}(x_{u,j,t})\cup\{\varnothing\}\bigr\}.
\end{aligned}
\label{eq:candidates}
\end{equation}
The symbol $\varnothing$ denotes direct use of the general procedure. Candidate pairs must satisfy skill input requirements, tool availability, and task constraints. The online routing policy selects an ordered sequence of skill-Adapter pairs from the valid candidates:
\begin{equation}
Z_{u,j,t}\sim\pi_{\mathrm{route}}\!\left(
\cdot\mid x_{u,j,t},\mathcal{C}_{u,j,t}
\right).
\label{eq:routing}
\end{equation}

Online routing combines user preferences, historical feedback, and the current execution state to select general-purpose skills and their associated Adapters. When the general procedure satisfies the current requirements, routing can use it directly. As task inputs or environment states change, routing can rematch the skill with an Adapter or select other general-purpose skills and their adapted versions.

\subsubsection{Incremental Maintenance of the Online Skill Library}

Routing results are written to the task-level online skill library $\mathcal{L}_{u,q,t}$. Each entry stores a reference to the general-purpose skill, the selected Adapter, applicable nodes, composition relations, and usage guidance. The library is updated as follows:
\begin{equation}
\begin{aligned}
\mathcal{L}_{u,q,t+1}
={}&\operatorname{Refresh}\!\bigl(
\mathcal{L}_{u,q,t},Z_{u,j,t},\\
&\quad E_{u,j,t}^{\mathrm{comp}},U_{u,j,t},x_{u,j,t}
\bigr).
\end{aligned}
\label{eq:refresh}
\end{equation}
Here, $E_{u,j,t}^{\mathrm{comp}}$ denotes valid input-output connections within the current skill composition, and $U_{u,j,t}$ denotes node-level usage guidance. During initialization, the routing agent creates entries for the first ready nodes in the DAG. The main LLM reads the skills and Adapters required by the current node and executes them, while the routing agent preloads candidates for subsequent nodes.

Based on execution feedback, the system reuses applicable entries, removes or replaces invalid entries, and retrieves skills and reselects Adapters for new requirements. For subsequent nodes that depend on upstream results, the system prepares candidates in advance, then checks applicability conditions and completes selection once the actual inputs are determined. The main LLM returns environment actions, execution observations, and the IDs of the skills and Adapters actually used, providing state information for the next routing round. After the task is completed, its trajectory and feedback support subsequent learning, and user Adapters can be reused in later tasks.

\subsection{Reinforcement Learning}

\subsubsection{Optimization Objective}

Personalized modification produces skill versions that satisfy specific requirements, while online routing determines the skills and Adapters to use at each step according to the execution state. We use GRPO to train $\pi_{\mathrm{mod}}$ and $\pi_{\mathrm{route}}$ separately. For the same input, we sample $K$ candidate decisions and compute group-relative advantages from their rewards:
\begin{equation}
\widehat{A}_k=
\frac{R_k-\operatorname{Mean}(\{R_l\}_{l=1}^{K})}
{\operatorname{Std}(\{R_l\}_{l=1}^{K})+\epsilon}.
\label{eq:advantage}
\end{equation}
Both policies are updated using the following objective:
\begin{equation}
\begin{aligned}
J(\theta)={}&\mathbb{E}\Biggl[
\frac{1}{K}\sum_{k=1}^{K}
\min\Bigl(\rho_k(\theta)\widehat{A}_k,\\
&\quad\operatorname{clip}\bigl(\rho_k(\theta),1-\epsilon_c,\\
&\qquad\quad 1+\epsilon_c\bigr)\widehat{A}_k\Bigr)\\
&\quad-\beta_{\mathrm{KL}}
D_{\mathrm{KL}}(\pi_\theta\Vert\pi_{\mathrm{ref}})
\Biggr].
\end{aligned}
\label{eq:grpo}
\end{equation}
Here, $\rho_k(\theta)$ is the ratio of the current policy's probability of a candidate decision to that of the old policy, $\epsilon_c$ is the clipping coefficient, and $\pi_{\mathrm{ref}}$ is the reference policy. The main LLM remains frozen during policy training.

\subsubsection{Reward for Personalized Modification}

The modification policy uses a set of trajectories to propose candidate Adapter updates and evaluates their effects on a held-out task set $\mathcal{Q}_{u}^{\mathrm{eval}}$. The general-purpose skill pool, user conditions, executor, and routing configuration are kept identical before and after modification. The modification reward is defined as the change in task success rate in percentage points:
\begin{equation}
\begin{aligned}
R^{\mathrm{mod}}=100\Bigl[&
\operatorname{SR}\!\bigl(
\mathcal{Q}_{u}^{\mathrm{eval}};\mathcal{G},\mathcal{A}_u'
\bigr)\\
&-\operatorname{SR}\!\bigl(
\mathcal{Q}_{u}^{\mathrm{eval}};\mathcal{G},\mathcal{A}_u
\bigr)
\Bigr].
\end{aligned}
\label{eq:mod-reward}
\end{equation}

The modification policy generates candidates from historical trajectories and evaluates execution outcomes using the requirements and feedback of the held-out tasks. Modifications that increase task success rate receive positive rewards, while those that decrease it receive negative rewards.

\subsubsection{Reward for Online Routing}

After training the modification policy, we train the routing policy with the current general-purpose skill pool and Adapter set. The routing action $Z_{u,j,t}$ is provided to the main LLM through the online skill library, and the task outcome $y_j$ is obtained after execution is complete. During training, a reference skill set $S_j^{\mathrm{gold}}$ measures the quality of general-purpose skill selection. Let
\begin{equation}
\begin{aligned}
S_j&=\operatorname{BaseIDs}(Z_{u,j,t}),\\
m_j&=|S_j\cap S_j^{\mathrm{gold}}|,\\
n_j&=|S_j|,\qquad k_j=|S_j^{\mathrm{gold}}|.
\end{aligned}
\label{eq:base-ids}
\end{equation}
The routing reward is
\begin{equation}
R_j^{\mathrm{route}}=
\begin{cases}
\displaystyle\frac{2m_j}{n_j+k_j},&y_j=\mathrm{success},\\[5pt]
-1,&y_j=\mathrm{failure}.
\end{cases}
\label{eq:route-reward}
\end{equation}

Successful tasks receive the F1 score between the selected general-purpose skills and the reference set, while failed tasks receive a negative reward. When both sets are empty, the F1 score for a successful task is defined as 1. Different Adapters share the ID of their parent general-purpose skill, so the effect of version selection enters the reward through the actual task outcome. This objective provides feedback on both task execution and skill selection to learn compact combinations of skills and Adapters suited to the current execution state. At execution time, the routing policy takes the current subtask, execution history, applicable requirements, and available candidates as inputs.

\section{Experiments}
\label{sec:experiments}

\subsection{Experimental Setup}

We evaluate Skill Adapter on ALFWorld, WebShop, and SkillsBench to examine how personalized modification and online routing of general-purpose skills affect task success and execution efficiency. We also analyze the effects of skill pool size, the organization of the online skill library, and reinforcement learning for the two policies.

\paragraph{Benchmarks and task conditions.}
ALFWorld (Shridhar et al.\ 2021) comprises six types of long-horizon tasks: object placement, cleaning, heating, cooling, examination, and manipulation of multiple objects. WebShop (Yao et al.\ 2022) requires agents to browse product pages and purchase items that satisfy user-specified attribute constraints. SkillsBench (Li et al.\ 2026b) evaluates agents' ability to identify, select, and use external skills. The three test sets contain 420, 1,500, and 210 tasks, respectively, with environment interaction budgets of 50, 15, and 30 steps. Personalization conditions are drawn from the goals, operational constraints, and output requirements explicitly stated in task instructions. Skill construction and policy training use execution experience from the training data.

\paragraph{Baselines.}
We compare seven configurations. Backbone (ReAct) (Yao et al.\ 2023) uses the main agent to execute tasks directly. Backbone (all skills) provides the main LLM with the entire skill pool. Backbone + BM25 retrieves skills through text matching (Robertson and Zaragoza 2009). LLM as Selector uses an LLM to select candidate skills. SkillRouter follows a retrieval-and-reranking pipeline (Zheng et al.\ 2026). Graph of Skills uses explicit skill relationships for structural retrieval (Li et al.\ 2026a). SkillRL optimizes skill use through reinforcement learning (Xia et al.\ 2026). Within each benchmark, all methods share the executor, initial general-purpose skill pool, task inputs, and interaction budget.

\paragraph{Metrics.}
For ALFWorld, we report success rates for the six task types, the overall success rate (SR), and mean task execution time. For WebShop, we report task score, success rate, and mean execution time. For SkillsBench, we report success rate, mean reward, and mean execution time. Execution time includes skill retrieval, modification and routing inference, main-LLM inference, and environment interaction.

\paragraph{Models and skill sources.}
We use Qwen3-8B (Yang et al.\ 2025) as the executor for ALFWorld and WebShop, and GPT-5-mini (OpenAI 2025) for SkillsBench. For the first two benchmarks, we cluster training trajectories from AgentTraj-L and use GPT-5.2-Codex to extract general-purpose skills. The initial SkillsBench skill pool is formed by merging the skills provided by the benchmark. The modification and routing policies use Qwen3-8B, with training inputs constructed from ALFWorld training trajectories and skills. The base procedures of general-purpose skills and their Adapters are stored separately, and routing results are supplied to the executor through the task-level online skill library.

\paragraph{Retrieval and training configuration.}
Candidate retrieval uses BGE-M3 embeddings. We first select three relevant topics and retrieve 20 skills, then expand the candidates along skill composition relationships and cap the candidate set at 30. The routing policy selects at most eight general-purpose skills. Both policies are trained with GRPO, using a group size of 8, a learning rate of $1\times10^{-6}$, a clipping coefficient of 0.2, a KL coefficient of 0.01, and a global batch size of 64. Each stage comprises 100 optimizer updates. The executor remains frozen during training. We use random seeds 42, 43, and 44, and report the mean results over three runs in the main tables.

\subsection{Main Results}

\subsubsection{Overall Performance}

Tables~\ref{tab:alfworld-webshop} and~\ref{tab:skillsbench} compare performance on the three benchmarks. Skill Adapter achieves the best results on all core metrics. Compared with SkillRL, success rates on ALFWorld, WebShop, and SkillsBench increase from 87.14\%, 72.73\%, and 21.43\% to 95.48\%, 81.53\%, and 38.57\%, respectively, corresponding to improvements of \textbf{8.34, 8.80, and 17.14 percentage points}.

\begin{table*}[t]
\centering
\begingroup
\fontsize{9}{10.5}\selectfont
\setlength{\tabcolsep}{3.5pt}
\renewcommand{\arraystretch}{1.15}
\begin{tabular}{@{}lrrrrrrrrrrr@{}}
\toprule
& \multicolumn{8}{c}{\textbf{ALFWorld}} & \multicolumn{3}{c}{\textbf{WebShop}} \\
\cmidrule(lr){2-9}\cmidrule(l){10-12}
\textbf{Method} & \textbf{Pick} & \textbf{Clean} & \textbf{Heat} & \textbf{Cool} & \textbf{Look} & \textbf{Pick 2} & \textbf{SR} $\uparrow$ & \textbf{Time (s)} $\downarrow$ & \textbf{Score} $\uparrow$ & \textbf{SR} $\uparrow$ & \textbf{Time (s)} $\downarrow$ \\
\midrule
Backbone (ReAct) & 59.05 & 60.49 & 39.58 & 42.67 & 48.72 & 56.94 & 52.86 & 88.6 & 46.00 & 19.80 & 72.4 \\
Backbone (all skills) & 30.48 & 33.33 & 20.83 & 25.33 & 28.21 & 23.61 & 27.62 & 121.5 & 24.80 & 7.53 & 103.7 \\
Backbone + BM25 & 68.57 & 61.73 & 56.25 & 56.00 & 53.85 & 65.28 & 61.67 & 81.3 & 58.13 & 25.27 & 66.8 \\
LLM as Selector & 70.48 & 67.90 & 60.42 & 56.00 & 61.54 & 68.06 & 65.00 & 99.7 & 61.00 & 28.87 & 82.5 \\
SkillRouter & 75.24 & 71.60 & 66.67 & 64.00 & 58.97 & 73.61 & 69.76 & 76.8 & 68.87 & 46.13 & 62.4 \\
Graph of Skills & 73.33 & 74.07 & 68.75 & 72.00 & 56.41 & 70.83 & 70.71 & 79.5 & 67.93 & 46.73 & 65.1 \\
SkillRL & \underline{94.29} & \underline{85.19} & \underline{85.42} & \underline{89.33} & \underline{69.23} & \underline{87.50} & \underline{87.14} & \underline{70.6} & \underline{85.20} & \underline{72.73} & \underline{57.8} \\
Skill Adapter (ours) & \textbf{98.10} & \textbf{95.06} & \textbf{95.83} & \textbf{94.67} & \textbf{89.74} & \textbf{95.83} & \textbf{95.48} & \textbf{61.9} & \textbf{92.67} & \textbf{81.53} & \textbf{50.7} \\
\bottomrule
\end{tabular}
\endgroup
\caption{Performance comparison on ALFWorld and WebShop with Qwen3-8B as the executor. The best results are boldfaced and the second-best results are underlined.}
\label{tab:alfworld-webshop}
\begin{minipage}{\textwidth}
\footnotesize Pick, Clean, Heat, Cool, Look, and Pick 2 denote the six ALFWorld task types. Higher success rates and scores are better; lower mean execution times are better.
\end{minipage}
\end{table*}

\begin{table}[tbp]
\centering
\begingroup
\fontsize{9}{10.5}\selectfont
\setlength{\tabcolsep}{3pt}
\renewcommand{\arraystretch}{1.15}
\begin{tabular*}{\linewidth}{@{\extracolsep{\fill}}lrrr@{}}
\toprule
\textbf{Method} & \textbf{SR} $\uparrow$ & \shortstack{\textbf{Mean}\\\textbf{reward} $\uparrow$} & \shortstack{\textbf{Mean time}\\\textbf{(s)} $\downarrow$} \\
\midrule
Backbone (ReAct) & 7.14 & 9.65 & 468.3 \\
Backbone (all skills) & 9.05 & 14.87 & 621.4 \\
Backbone + BM25 & 13.33 & 17.24 & 525.2 \\
LLM as Selector & 14.76 & 18.63 & 587.3 \\
SkillRouter & 17.62 & 19.46 & 615.0 \\
Graph of Skills & 18.57 & 23.09 & \underline{462.9} \\
SkillRL & \underline{21.43} & \underline{25.33} & 529.4 \\
Skill Adapter (ours) & \textbf{38.57} & \textbf{34.90} & \textbf{398.9} \\
\bottomrule
\end{tabular*}
\endgroup
\caption{Performance comparison on SkillsBench with GPT-5-mini as the executor. The best results are boldfaced and the second-best results are underlined.}
\label{tab:skillsbench}
\end{table}

Skill Adapter achieves the highest success rate on all six ALFWorld task types. Compared with SkillRL, the success rate on Look tasks increases from 69.23\% to 89.74\%, while gains on the other five task types range from 3.81 to 10.41 percentage points. The WebShop task score increases from 85.20 to 92.67, and the SkillsBench mean reward increases from 25.33 to 34.90.

On all three benchmarks, configurations that select skills achieve higher task success rates than the configuration that provides the main LLM with the entire skill pool.

\subsubsection{Online Routing and Execution Efficiency}

The main LLM reads the skills and Adapters available for the current node from the online skill library, while the routing agent preloads candidates for subsequent nodes along the DAG. When subgoals, intermediate results, or environment feedback change the applicability conditions, routing retains entries that remain applicable and updates the affected skills and their versions.

Compared with SkillRL, Skill Adapter reduces mean execution time from 70.6 to 61.9 seconds on ALFWorld, from 57.8 to 50.7 seconds on WebShop, and from 529.4 to 398.9 seconds on SkillsBench.

\subsubsection{Scaling the Skill Pool}

\begin{figure}[t]
\centering
\includegraphics[width=\linewidth]{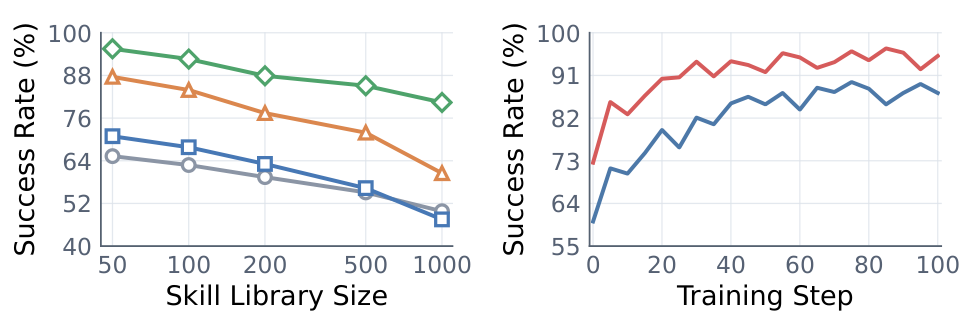}
\caption{Skill pool scaling and policy training. Left: success rates at different skill pool sizes. Right: changes in the success rates of the two policies over 100 training steps.}
\label{fig:scaling-training}
\end{figure}

The scaling experiment starts with the same task-relevant skills and gradually adds distractor skills from other task categories. The left panel of Figure~\ref{fig:scaling-training} reports success rates as the skill pool expands from 50 to 1,000 skills.

Success rates decline for all methods as the skill pool grows. Skill Adapter decreases from approximately 95.5\% to 80.4\%, while SkillRL decreases from approximately 87.2\% to 60.3\%. At 1,000 skills, Skill Adapter retains a lead of approximately 20 percentage points. Compared with BM25, Graph of Skills achieves success rates that are 9.04, 21.46, and 5.24 percentage points higher on the three benchmarks, respectively; Skill Adapter further exceeds Graph of Skills by 24.77, 34.80, and 20.00 percentage points.

\subsection{Ablation Study}

Table~\ref{tab:ablation} analyzes the effects of general-purpose skill representations, the organization of online routing, and training the modification and routing policies.

\begin{table}[htbp]
\centering
\begingroup
\fontsize{9}{10.5}\selectfont
\setlength{\tabcolsep}{2pt}
\renewcommand{\arraystretch}{1.2}
\begin{tabular}{@{}l>{\raggedright\arraybackslash}p{0.30\linewidth}rrrr@{}}
\toprule
\textbf{Group} & \textbf{Variant} & \textbf{ALF.} $\uparrow$ & \textbf{Web.} $\uparrow$ & \textbf{SB.} $\uparrow$ & \textbf{Avg.} $\uparrow$ \\
\midrule
Skills & Trajectory pool only & 52.86 & 25.93 & 9.05 & 29.28 \\
Skills & + General-purpose skill pool & 70.24 & 41.53 & 13.81 & 41.86 \\
Routing & + Composition and usage guidance & 76.90 & 51.47 & 16.67 & 48.35 \\
Routing & + DAG preloading & 80.71 & 57.07 & 20.48 & 52.75 \\
RL & Modification policy & 87.86 & 68.80 & 23.81 & 60.16 \\
RL & Routing policy & \underline{90.95} & \underline{73.93} & \underline{30.95} & \underline{65.28} \\
Full & Skill Adapter & \textbf{95.48} & \textbf{81.53} & \textbf{38.57} & \textbf{71.86} \\
\bottomrule
\end{tabular}
\endgroup
\caption{Ablation study of general-purpose skills, online routing, and reinforcement learning (RL). ALF., Web., and SB. denote success rates on the three benchmarks, respectively; Avg. is their arithmetic mean.}
\label{tab:ablation}
\end{table}

\subsubsection{General-Purpose Skills and Online Routing Organization}

Adding a general-purpose skill pool to the trajectory pool increases mean success rate from 29.28\% to 41.86\%, an improvement of 12.58 percentage points. Adding skill composition relationships and usage guidance raises mean success rate to 48.35\%. Further adding DAG preloading increases it to 52.75\%, a gain of 4.40 percentage points.

\subsubsection{Training the Modification and Routing Policies}

The configuration with reinforcement learning for the modification policy achieves a mean success rate of 60.16\%, while the configuration with reinforcement learning for the routing policy achieves 65.28\%. The full method uses both trained policies and achieves a mean success rate of 71.86\%, exceeding training the routing policy alone by 6.58 percentage points.

In the right panel of Figure~\ref{fig:scaling-training}, the success rates of the modification and routing policies increase from 59.5\% and 72\% to approximately 87\% and 95\%, respectively.

\section{Conclusion}
\label{sec:conclusion}

Skill Adapter stores users' specific requirements for tools, operating procedures, and outputs as Adapters linked to general-purpose skills. As a task progresses, the routing agent selects general-purpose skills and applicable Adapters, incrementally maintains an online skill library that the main LLM reads on demand, and preloads candidates for subsequent nodes. GRPO trains the two policies using modification gains and routing outcomes.

On ALFWorld, WebShop, and SkillsBench, Skill Adapter improves success rates over SkillRL by an average of 11.43 percentage points and reduces mean execution time on all three benchmarks. In the skill pool scaling experiment, Skill Adapter achieves a higher success rate than SkillRL in the setting with 1,000 skills.

\clearpage
\section*{References}
\phantomsection
\addcontentsline{toc}{section}{References}
\begingroup
\sloppy
\normalsize
\begin{list}{}{\setlength{\leftmargin}{0pt}\setlength{\labelwidth}{0pt}\setlength{\itemindent}{0pt}\setlength{\itemsep}{2.8pt}\setlength{\parsep}{0pt}}
\item Ahn, M.; Brohan, A.; Brown, N.; Chebotar, Y.; Cortes, O.; David, B.; Finn, C.; Fu, C.; Gopalakrishnan, K.; Hausman, K.; et al. 2022. Do As I Can, Not As I Say: Grounding Language in Robotic Affordances. arXiv:2204.01691.

\item Bai, T.; Wan, Z.; Zhou, P.; Yu, X.; Zhao, W.; You, Y.; and Tsang, I. W. 2026. SkillDAG: Self-Evolving Typed Skill Graphs for LLM Skill Selection at Scale. arXiv:2606.03056.

\item Deng, X.; Gu, Y.; Zheng, B.; Chen, S.; Stevens, S.; Wang, B.; Sun, H.; and Su, Y. 2023. Mind2Web: Towards a Generalist Agent for the Web. arXiv:2306.06070.

\item Li, D.; Li, Z.; Du, H.; Wu, X.; Gui, S.; Kuang, Y.; and Sun, L. 2026a. Graph of Skills: Dependency-Aware Structural Retrieval for Massive Agent Skills. arXiv:2604.05333.

\item Li, X.; Chen, W.; Liu, Y.; Zheng, S.; Chen, X.; He, Y.; Li, Y.; et al. 2026b. SkillsBench: Benchmarking How Well Agent Skills Work Across Diverse Tasks. arXiv:2602.12670.

\item Liang, Y.; Zhong, R.; Xu, H.; Jiang, C.; Zhong, Y.; Fang, R.; Gu, J.-C.; et al. 2026. SkillNet: Create, Evaluate, and Connect AI Skills. arXiv:2603.04448.

\item Liu, X.; Yu, H.; Zhang, H.; Xu, Y.; Lei, X.; Lai, H.; Gu, Y.; Ding, H.; Men, K.; Yang, K.; et al. 2023. AgentBench: Evaluating LLMs as Agents. arXiv:2308.03688.

\item Liu, Y.; Ji, J.; An, L.; Jaakkola, T.; Zhang, Y.; and Chang, S. 2026. How Well Do Agentic Skills Work in the Wild: Benchmarking LLM Skill Usage in Realistic Settings. arXiv:2604.04323.

\item Lu, Y.; Yu, H.; and Khashabi, D. 2023. GEAR: Augmenting Language Models with Generalizable and Efficient Tool Resolution. arXiv:2307.08775.

\item OpenAI. 2025. GPT-5 System Card. \url{https://openai.com/index/gpt-5-system-card/}. Accessed July 20, 2026.

\item Packer, C.; Wooders, S.; Lin, K.; Fang, V.; Patil, S. G.; Stoica, I.; and Gonzalez, J. E. 2023. MemGPT: Towards LLMs as Operating Systems. arXiv:2310.08560.

\item Paranjape, B.; Lundberg, S.; Singh, S.; Hajishirzi, H.; Zettlemoyer, L.; and Ribeiro, M. T. 2023. ART: Automatic Multi-step Reasoning and Tool-use for Large Language Models. arXiv:2303.09014.

\item Park, J. S.; O'Brien, J. C.; Cai, C. J.; Morris, M. R.; Liang, P.; and Bernstein, M. S. 2023. Generative Agents: Interactive Simulacra of Human Behavior. In \emph{Proceedings of the 36th Annual ACM Symposium on User Interface Software and Technology}, 1--22.

\item Patil, S. G.; Zhang, T.; Wang, X.; and Gonzalez, J. E. 2023. Gorilla: Large Language Model Connected with Massive APIs. arXiv:2305.15334.

\item Qian, C.; Han, C.; Fung, Y. R.; Qin, Y.; Liu, Z.; and Ji, H. 2023. CREATOR: Tool Creation for Disentangling Abstract and Concrete Reasoning of Large Language Models. arXiv:2305.14318.

\item Qin, Y.; Liang, S.; Ye, Y.; Zhu, K.; Yan, L.; Lu, Y.; Lin, Y.; Cong, X.; Tang, X.; Qian, B.; et al. 2023. ToolLLM: Facilitating Large Language Models to Master 16000+ Real-world APIs. arXiv:2307.16789.

\item Robertson, S.; and Zaragoza, H. 2009. The Probabilistic Relevance Framework: BM25 and Beyond. \emph{Foundations and Trends in Information Retrieval}, 3(4): 333--389.

\item Schick, T.; Dwivedi-Yu, J.; Dessì, R.; Raileanu, R.; Lomeli, M.; Zettlemoyer, L.; Cancedda, N.; and Scialom, T. 2023. Toolformer: Language Models Can Teach Themselves to Use Tools. arXiv:2302.04761.

\item Schulman, J.; Wolski, F.; Dhariwal, P.; Radford, A.; and Klimov, O. 2017. Proximal Policy Optimization Algorithms. arXiv:1707.06347.

\item Shao, Z.; Wang, P.; Zhu, Q.; Xu, R.; Song, J.; Bi, X.; Zhang, H.; et al. 2024. DeepSeekMath: Pushing the Limits of Mathematical Reasoning in Open Language Models. arXiv:2402.03300.

\item Shen, J.; Zhang, T.; Zhao, X.; and Cheng, H. 2026. Dynamic Skill Lifecycle Management for Agentic Reinforcement Learning. arXiv:2605.10923.

\item Shinn, N.; Cassano, F.; Berman, E.; Gopinath, A.; Narasimhan, K.; and Yao, S. 2023. Reflexion: Language Agents with Verbal Reinforcement Learning. In \emph{Advances in Neural Information Processing Systems}.

\item Shridhar, M.; Yuan, X.; Côté, M.-A.; Bisk, Y.; Trischler, A.; and Hausknecht, M. 2021. ALFWorld: Aligning Text and Embodied Environments for Interactive Learning. In \emph{International Conference on Learning Representations}.

\item Tu, S.; Xu, C.; Zhang, Q.; Zhang, Y.; Lan, X.; Li, L.; Li, D.; and Zhao, D. 2026. Dynamic Dual-Granularity Skill Bank for Agentic RL. arXiv:2603.28716.

\item Wang, G.; Xie, Y.; Jiang, Y.; Mandlekar, A.; Xiao, C.; Zhu, Y.; Fan, L.; and Anandkumar, A. 2023. Voyager: An Open-Ended Embodied Agent with Large Language Models. arXiv:2305.16291.

\item Wang, J.; Yan, Q.; Wang, Y.; Tian, Y.; Mishra, S. S.; Xu, Z.; Gandhi, M.; Xu, P.; and Cheong, L. L. 2025. Reinforcement Learning for Self-Improving Agent with Skill Library. arXiv:2512.17102.

\item Wu, Q.; Bansal, G.; Zhang, J.; Wu, Y.; Li, B.; Zhu, E.; Jiang, L.; Zhang, X.; Zhang, S.; Liu, J.; Awadallah, A. H.; White, R. W.; Burger, D.; and Wang, C. 2023. AutoGen: Enabling Next-Gen LLM Applications via Multi-Agent Conversation. arXiv:2308.08155.

\item Xia, P.; Chen, J.; Wang, H.; Liu, J.; Zeng, K.; Wang, Y.; Han, S.; et al. 2026. SkillRL: Evolving Agents via Recursive Skill-Augmented Reinforcement Learning. arXiv:2602.08234.

\item Xie, T.; Zhang, D.; Chen, J.; Li, X.; Zhao, S.; Cao, R.; Hua, T. J.; Cheng, Z.; Shin, D.; Lei, F.; Liu, Y.; Xu, Y.; Zhou, S.; Savarese, S.; Xiong, C.; Zhong, V.; and Yu, T. 2024. OSWorld: Benchmarking Multimodal Agents for Open-Ended Tasks in Real Computer Environments. In \emph{Advances in Neural Information Processing Systems}, volume 37.

\item Yang, A.; Li, A.; Yang, B.; Zhang, B.; Hui, B.; Zheng, B.; Yu, B.; et al. 2025. Qwen3 Technical Report. arXiv:2505.09388.

\item Yang, J.; Jimenez, C. E.; Wettig, A.; Lieret, K.; Yao, S.; Narasimhan, K.; and Press, O. 2024. SWE-agent: Agent-Computer Interfaces Enable Automated Software Engineering. arXiv:2405.15793.

\item Yao, S.; Chen, H.; Yang, J.; and Narasimhan, K. 2022. WebShop: Towards Scalable Real-World Web Interaction with Grounded Language Agents. In \emph{Advances in Neural Information Processing Systems}.

\item Yao, S.; Zhao, J.; Yu, D.; Du, N.; Shafran, I.; Narasimhan, K.; and Cao, Y. 2023. ReAct: Synergizing Reasoning and Acting in Language Models. In \emph{International Conference on Learning Representations}.

\item Zhao, A.; Huang, D.; Xu, Q.; Lin, M.; Liu, Y.-J.; and Huang, G. 2023. ExpeL: LLM Agents Are Experiential Learners. arXiv:2308.10144.

\item Zheng, Y.; Zhang, Z.; Ma, C.; Yu, Y.; Zhu, J.; Dong, B.; and Zhu, H. 2026. SkillRouter: Retrieve-and-Rerank Skill Selection for LLM Agents at Scale. arXiv:2603.22455.

\item Zhou, S.; Xu, F. F.; Zhu, H.; Zhou, X.; Lo, R.; Sridhar, A.; Cheng, X.; Ou, T.; Bisk, Y.; Fried, D.; Alon, U.; and Neubig, G. 2024. WebArena: A Realistic Web Environment for Building Autonomous Agents. In \emph{International Conference on Learning Representations}.
\end{list}
\endgroup

\end{document}